\documentclass{article}

\usepackage[final]{corl_2026} % Uncomment for the camera-ready ``final'' version.

\usepackage{graphicx}
\usepackage{amsmath}
\usepackage{amssymb}
\usepackage{tabularx}
\usepackage{multirow}
\usepackage{booktabs}
\usepackage{enumitem}
\usepackage{subcaption}
\newcommand{\myparagraph}[1]{\noindent\textbf{#1}}

\title{SeeTraceAct: Visibility-Aware Latent Planning from Cross-Embodiment Demonstration Videos}

\author{
  \textbf{Jaehyeon Son}$^{1,2}$
  \quad \textbf{Junhyun Kim}$^1$
  \quad \textbf{Kyle Kam}$^1$
  \quad \textbf{Jeremiah Coholich}$^1$
  \quad \textbf{Seok Joon Kim}$^1$ \\[0.2em]
  \textbf{Jinhoo Kim}$^1$
  \quad \textbf{Chris Dongjoo Kim}$^2$
  \quad \textbf{Jaemin Cho}$^{2,3}$
  \quad \textbf{Dieter Fox}$^{2,4}$
  \quad \textbf{Zsolt Kira}$^1$ \\[0.4em]
  $^1$Georgia Institute of Technology
  \quad $^2$Allen Institute for AI \\
  $^3$Johns Hopkins University
  \quad $^4$University of Washington
}

\begin{document}
\maketitle
\vspace{-3.1em}

%===============================================================================

\begin{center}
  \url{https://github.com/jaehyeon-son/SeeTraceAct}
\end{center}

\vspace{0.7em}

\begin{abstract}
  Vision-language-action models (VLAs) are promising general-purpose robot policies, but adapting them to new tasks typically requires costly task-specific teleoperation data.
  As an alternative, we study one-shot demo-conditioned VLAs, where a robot policy is conditioned on a single demonstration video of an unseen task.
  We find that existing end-to-end approaches often struggle when successful execution requires precisely localizing small target regions.
  To address this limitation, we propose \textsc{SeeTraceAct}, a demo-conditioned VLA framework that encourages precise spatial grounding through visibility-aware prediction of future end-effector traces.
  To enable reproducible evaluation with cross-embodiment demonstrations, we introduce and release \textsc{RoboCasa-DC}, a demo-conditioned extension of RoboCasa that pairs its robot trajectories with newly collected humanoid demonstrations.
  Experiments on \textsc{RoboCasa-DC} and a real-world benchmark, where a Franka Panda arm is conditioned on human demonstrations, show that \textsc{SeeTraceAct} outperforms baselines, achieving the best success rate across all four \textsc{RoboCasa-DC} settings and improving real-world average success by 9.5 percentage points.
\end{abstract}

% \keywords{Vision-Language-Action Models, Cross-Embodiment Learning}

%===============================================================================

\section{Introduction}
\label{sec:intro}

\begin{figure}[t]
	\centering
	\includegraphics[width=\textwidth,trim=0 215 0 6,clip]{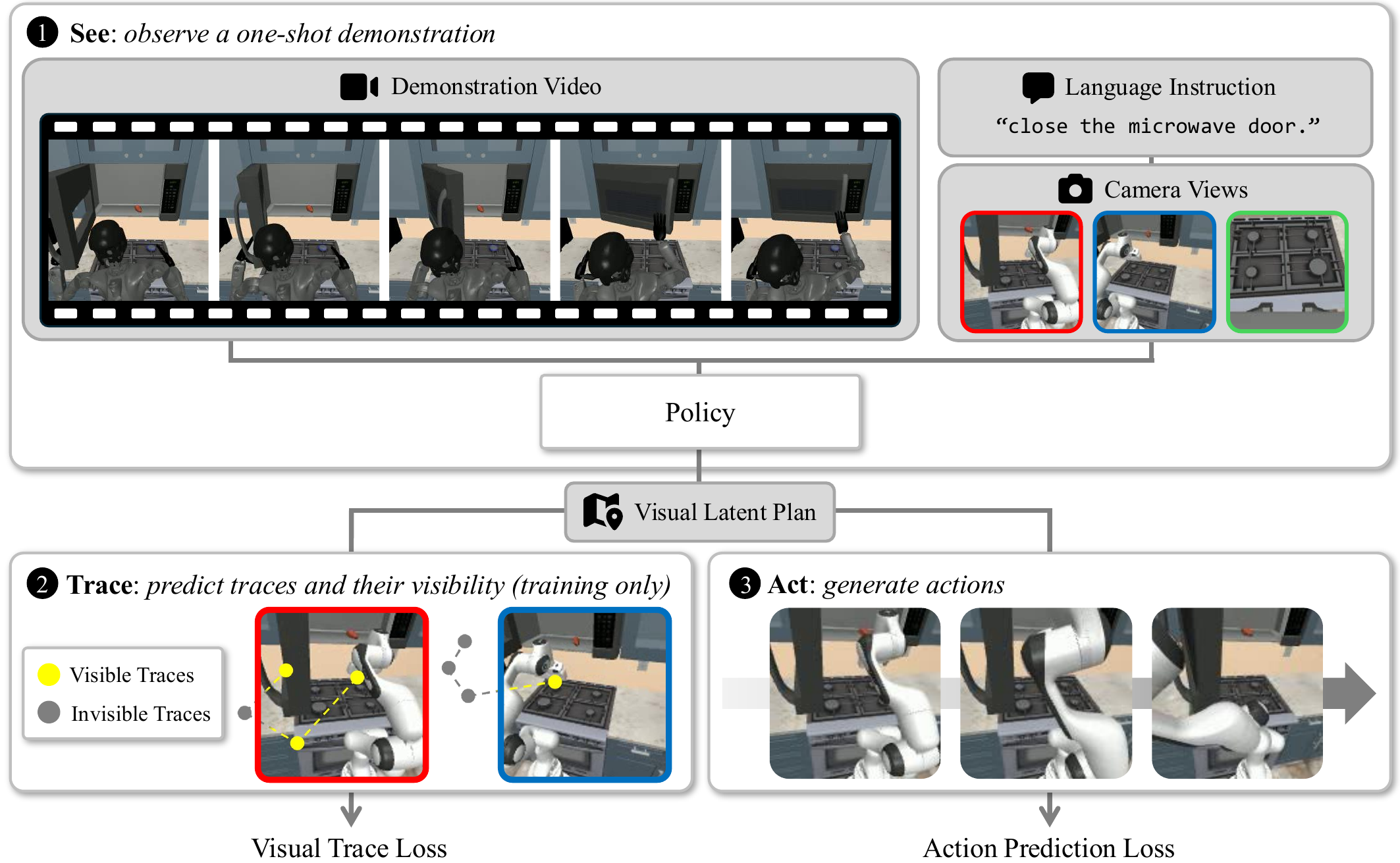}
	\caption{Overview of \textsc{SeeTraceAct}.
  Given a demonstration video, current camera views, and a language instruction, the policy encodes task-relevant information into a \emph{visual latent plan} (\textsc{See}).
  During training, the policy predicts future visual traces and their visibility for each camera view (\textsc{Trace}), while also predicting actions from the latent plan (\textsc{Act}).
  At inference time, the trace prediction component is discarded, and the policy generates actions from the latent plan (\textsc{Act}).
	}
	\label{fig:overview}
\vskip -0.1in
\end{figure}

Vision-language-action models (VLAs) have recently emerged as a promising path toward general-purpose robotic agents by adapting pretrained vision-language models (VLMs) for robotic control~\citep{pi0, OpenVLA, GR00T}.
However, deploying VLAs on new tasks typically requires post-training on expert robot trajectories collected through teleoperation.
Because collecting teleoperated robot data requires specialized hardware and substantial human effort, it remains a central bottleneck for scaling VLAs across diverse tasks, embodiments, and environments~\citep{BridgeDataV2, DROID, OXE}.

A scalable alternative is to let an end user specify a new robot task by performing it once in front of a camera.
This motivates one-shot demo-conditioned VLAs, where a policy is conditioned on a single human demonstration video of an unseen target task.
However, this problem is non-trivial: the cross-embodiment gap makes directly translating human demonstrations into robot actions fundamentally challenging.

While methods that utilize explicit embodiment-specific retargeting have been developed~\citep{Demodiffusion, OKAMI, DITTO}, they require embodiment- and task-specific expertise beyond what an end user can provide, so we instead focus on end-to-end approaches.
Prior work in this direction has learned video representations and employed auxiliary objectives to ground demonstrations~\citep{Vid2Robot, Uniskill, ViVLA}.
While these methods have shown promise, we find that they often struggle with precision-sensitive tasks, such as pressing the target button on a coffee machine or turning on a sink faucet (see \S\ref{sec:exp:contrast}).

To address this limitation, we propose \textsc{SeeTraceAct}, a demo-conditioned VLA framework that encourages precise spatial grounding through an auxiliary future-trace prediction objective (Fig.~\ref{fig:overview}).
Given a demonstration video, \textsc{SeeTraceAct} learns a \emph{visual latent plan} that summarizes the task-relevant motion the robot should execute.
During training, this latent plan is decoded into future end-effector traces in the robot's camera views.
However, trace supervision is not always well-defined, especially in multi-view setups: the end effector may move off-screen in some camera views, making its image coordinates ill-posed targets.
\textsc{SeeTraceAct} therefore uses a visibility-aware trace decoder that predicts both trace coordinates and their validity, preserving supervision even when some trace points are off-screen.
At inference time, the trace decoder is discarded, and the policy generates actions directly from the latent plan.

To support reproducible evaluation and future work on demo-conditioned VLAs, we introduce and release the \textsc{RoboCasa-DC} benchmark, a demo-conditioned extension of the RoboCasa~\citep{RoboCasa} simulation environment (\S\ref{sec:robocasa-dc}).
\textsc{RoboCasa-DC} supports both same-embodiment and cross-embodiment evaluation, allowing policies to be conditioned on demonstration videos from either the Panda arm itself or a humanoid that serves as a proxy for human demonstrators.
This complements prior evaluations that focus on real-world settings~\citep{Vid2Robot}, use single-arm robot demonstrations in simulation~\citep{ViVLA}, or do not release the human demonstrations used for simulated evaluation~\citep{Uniskill}.

Experiments on \textsc{RoboCasa-DC} and a real-world benchmark with a Franka Panda arm show that \textsc{SeeTraceAct} achieves the strongest results among competitive demo-conditioned VLA baselines (\S\ref{sec:exp:results}).
On \textsc{RoboCasa-DC}, \textsc{SeeTraceAct} achieves the best success rate across all four evaluation settings.
On the real-world benchmark, where the robot is conditioned on human demonstrations, \textsc{SeeTraceAct} improves the average success rate by 9.5 percentage points.
Our ablation study (\S\ref{sec:exp:ablation}) supports the importance of our key design choices, including visibility-aware trace supervision.
Our main contributions are as follows:
\begin{enumerate}[leftmargin=16pt, itemsep=2.5pt, topsep=2.5pt, parsep=1pt]
  \item We introduce \textsc{SeeTraceAct}, a demo-conditioned VLA framework that encourages precise spatial grounding of one-shot demonstration videos, including cross-embodiment ones, via visibility-aware prediction of future end-effector traces.
  \item We introduce and release \textsc{RoboCasa-DC}, a demo-conditioned extension of RoboCasa for benchmarking policies with both same- and cross-embodiment demonstration videos.
  \item We show that \textsc{SeeTraceAct} achieves the strongest results among competitive baselines on \textsc{RoboCasa-DC} and a real-world benchmark, with ablations supporting our key design choices.
\end{enumerate}
\vspace{-1pt}

\section{Related Work}
\label{sec:rel}

\myparagraph{Vision-language-action models (VLAs).}
Early large-scale VLAs~\citep{RT1, RT2} introduce transformer-based policies trained on extensive robot trajectories paired with web-scale vision-language data.
OpenVLA~\citep{OpenVLA} open-sources a VLA pretrained on the Open X-Embodiment dataset~\citep{OXE}.
More recently, $\pi_0$~\citep{pi0} and GR00T N1~\citep{GR00T} combine a pretrained vision-language backbone with a diffusion-based action expert for continuous, high-frequency control.
However, these VLAs still require substantial task- and embodiment-specific teleoperation data for post-training, motivating our focus on one-shot demonstration videos.

\myparagraph{One-shot demo-conditioned policies.}
One line of work uses kinematic retargeting between humans and robots to learn from one-shot demonstrations~\citep{Demodiffusion, OKAMI, DITTO}.
These approaches enable explicit cross-embodiment transfer from human demonstrations to robot actions, but often rely on substantial task- or embodiment-specific engineering.
We instead focus on end-to-end approaches that condition robot policies directly on demonstrations without explicit retargeting.
XSkill~\citep{xskill} and UniSkill~\citep{Uniskill} learn cross-embodiment skill representations from human and robot videos, while Vid2Robot~\citep{Vid2Robot} trains an end-to-end demo-conditioned policy with video- and language-alignment objectives.
ViVLA~\citep{ViVLA} learns a unified latent action space across human and robot demonstrations and trains a VLA to predict both latent action tokens and robot actions.
While these methods have shown promise, we find that they can still struggle when task success depends on precisely localizing small interaction regions (\S\ref{sec:exp:contrast}).
\textsc{SeeTraceAct} addresses this limitation through an auxiliary visual trace objective that encourages precise spatial grounding.

\myparagraph{Visual traces in VLAs.}
Several works have incorporated visual traces into VLAs:
LLARVA~\citep{LLARVA} and MolmoAct~\citep{MolmoAct} predict future visual traces and feed them autoregressively to the action model, while RT-Trajectory~\citep{RT-Trajectory} uses trajectory sketches obtained from human drawings, videos, or model predictions.
HAMSTER~\citep{HAMSTER} uses 2D traces as an intermediate representation in hierarchical policies, and TraceVLA~\citep{TraceVLA} conditions the policy on past traces extracted from previous observations.
Closer to our approach, ThinkAct~\citep{ThinkAct} and FastThinkAct~\citep{FastThinkAct} condition action generation on visual latent plans rather than decoded visual traces.
\textsc{SeeTraceAct} differs by using visual traces to ground demonstration videos into spatial plans while explicitly handling off-screen trace points.

\section{Method}
\label{sec:appr}

\subsection{Problem setting}

We consider a setting in which a policy is evaluated on unseen tasks, conditioned on a single demonstration video that may come from a different embodiment.
Let $\mathcal{T}$ denote a distribution over robotic tasks, and let $\mathcal{T}_{\mathrm{seen}}$ and $\mathcal{T}_{\mathrm{unseen}}$ denote disjoint sets of seen and held-out unseen tasks sampled from this distribution.
During training, for each seen task $\kappa \in \mathcal{T}_{\mathrm{seen}}$, the policy $\pi_{\theta}$ has access to multiple training tuples $(\xi^{\kappa}, D^{\kappa}, l^{\kappa})$, where $\xi^{\kappa} = \{(o_t, q_t, a_t)\}_{t=1}^{T}$ is an expert trajectory from the policy embodiment, $D^{\kappa}$ is a demonstration video that may come from a different embodiment, and $l^{\kappa}$ is the corresponding language instruction.
Here, $o_t$, $q_t$, and $a_t$ denote the camera view, robot state, and robot action at timestep $t$, respectively.
We omit the task superscript for brevity.

At evaluation, the policy is given a single demonstration video $D$ and language instruction $l$ for an unseen task $\kappa^\ast \in \mathcal{T}_{\mathrm{unseen}}$.
At each timestep $t$, it takes the current observation $(o_t, q_t)$ and predicts an action chunk $A_t = \{a_t, \dots, a_{t+H-1}\}$ toward completing $\kappa^\ast$, where $H$ denotes the action horizon:
$ A_t \sim \pi_{\theta}(\cdot \mid o_t, q_t, l, D). $

\begin{figure}[t]
	\centering
	\includegraphics[width=\textwidth,trim=2 410 0 1,clip]{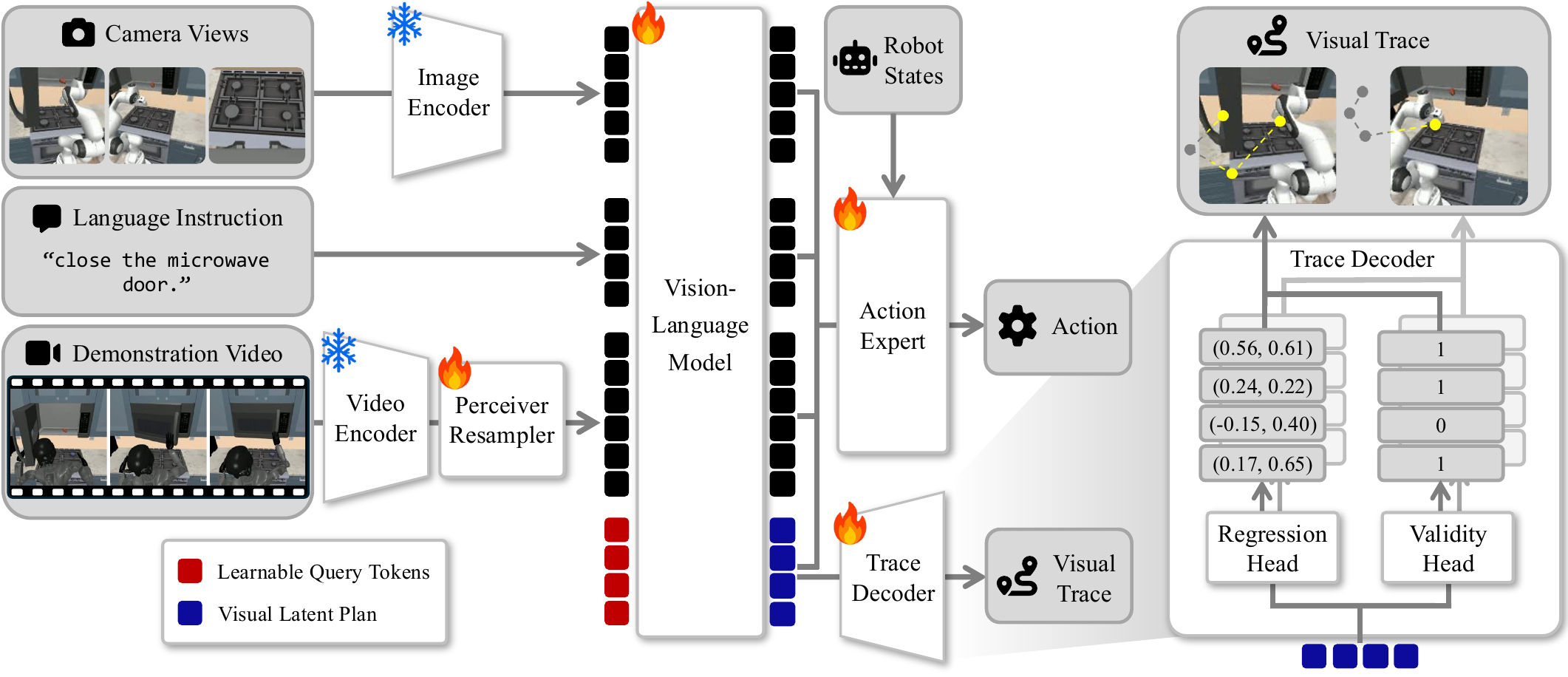}
	\caption{Architecture of \textsc{SeeTraceAct}.
	It receives camera views, a language instruction, a demonstration video, and robot states, and outputs an action chunk.
	We append learnable query tokens after the input tokens;
  their final hidden states form a \textit{visual latent plan}, which is decoded into future end-effector traces during training.
	The trace decoder consists of a regression head that predicts the trace coordinates and a validity head that predicts whether each trace point is in-frame.
	The trace decoder is used only during training and discarded at inference time.
	}
	\label{fig:arch}
  \vskip -0.15in
\end{figure}

\subsection{Architecture}

We build \textsc{SeeTraceAct} on top of the GR00T N1.5~\citep{GR00T} architecture, which consists of a vision-language model (VLM) and a flow-matching-based action expert.
As shown in Fig.~\ref{fig:arch}, our model augments this architecture with video encoding modules, learnable query tokens that form a visual latent plan, and a trace decoder.
At each control timestep $t$, the VLM processes image and language tokens from the current camera views and language instruction, producing a representation $\phi_t$.

\myparagraph{Video tokens.}
The left side of Fig.~\ref{fig:arch} shows how we incorporate the demonstration video.
We first encode the demonstration $D$ using a video encoder~\citep{V-JEPA-2} pre-trained on large-scale video data, including action-centric datasets~\citep{SSV2, Kinetics}.
We add a Perceiver Resampler~\citep{Flamingo,Perceiver} on top of the video encoder to compress the extracted features into a compact set of video tokens (from $8{,}192$ to $32$ tokens; see App.~\ref{sec:model-det}), and append them to the image and language tokens in the VLM input sequence.
Under causal attention, this ordering lets the video tokens attend to the image and language tokens, making the demonstration representation context-dependent rather than fixed across the episode.
We denote the final VLM hidden states corresponding to these video tokens by $\psi_t$.

\myparagraph{Visual latent plan.}
The middle of Fig.~\ref{fig:arch} illustrates how we form the visual latent plan.
Following recent visual latent planning work~\citep{ThinkAct, FastThinkAct}, we use this term to refer to a latent representation that is trained to encode future task progression and conditions action generation.
Rather than assigning this planning role directly to the existing input tokens, we append a separate set of learnable query tokens after them.
Under causal attention, these query tokens attend to the image, language, and video tokens, and their final hidden states constitute the visual latent plan $z_t$.

\myparagraph{Visibility-aware trace decoder.}
As shown on the right side of Fig.~\ref{fig:arch}, the trace decoder predicts future end-effector traces in each static camera view from the visual latent plan $z_t$.
However, the end effector may move off-screen in some camera views, making its normalized 2D image coordinates ill-posed regression targets.
To address this, we use a \textit{visibility-aware} trace decoder with two heads.
The regression head predicts normalized end-effector coordinates, while the validity head predicts whether or not each trace point is in-frame, allowing off-screen points to still provide a learning signal (see ablation in \S\ref{sec:exp:ablation}).
The trace decoder is discarded after training.

\subsection{Training and inference}

We train \textsc{SeeTraceAct} with two objectives: the \textit{action prediction loss} and the auxiliary \textit{visual trace loss}.
The former optimizes the policy to predict robot action chunks, while the latter shapes the visual latent plan $z_t$ to encode future task progression in the robot's image space.

\myparagraph{Action prediction loss.}
Following prior works~\citep{pi0,GR00T}, we train the action expert with an action prediction loss instantiated as a flow-matching objective~\citep{FlowMatching}.
At each control timestep $t$, we sample noise $\epsilon \sim \mathcal{N}(\mathbf{0}, \mathbf{I})$ and draw a flow-matching timestep $\tau \in [0,1]$ from a shifted beta distribution.
Given the ground-truth action chunk $A_t$, we construct the noised action chunk as
$ A_t^\tau = \tau A_t + (1-\tau)\epsilon. $
The action prediction loss trains the action expert $V_\theta$ to predict the target velocity field from the VLM outputs $\phi_t$, $\psi_t$, and $z_t$, together with $q_t$ and the noised action chunk $A_t^\tau$:
\begin{equation}
\mathcal{L}_{\mathrm{act}}(\theta)
=
\mathbb{E}_{\tau, \epsilon}
\left[
\left\|
V_{\theta}(\phi_t, \psi_t, z_t, A_t^{\tau}, q_t)
-
(A_t - \epsilon)
\right\|^2
\right].
\label{eq:loss_act}
\end{equation}

\myparagraph{Visual trace loss.}
We supervise the policy with visibility-aware 2D image-space traces, directly aligning the target with the robot's visual observation space.
For each camera view, we predict $N$ future trace points at a temporal stride $\Delta$ using $N$ learnable query tokens.
We denote the final hidden states of these tokens by
$ z_t = \{z_{t,1}, \dots, z_{t,N}\}. $
The corresponding trace target is
$ Y_t = \{y_{t+\Delta}, y_{t+2\Delta}, \dots, y_{t+N\Delta}\}, $
where each $z_{t,n}$ is supervised to predict $y_{t+n\Delta}$, the relative coordinate of the end effector.
We also define the binary validity target
$ M_t = \{m_{t+\Delta}, m_{t+2\Delta}, \dots, m_{t+N\Delta}\}, $
where $m_{t+n\Delta}$ indicates whether the end effector is in-frame.
The trace decoder takes $z_t$ as input and applies two heads.
The regression head outputs the predicted trace
$
\hat{Y}_t = \{\hat{y}_{t+\Delta}, \dots, \hat{y}_{t+N\Delta}\},
$
while the validity head outputs the predicted validity scores
$
\hat{M}_t = \{\hat{m}_{t+\Delta}, \dots, \hat{m}_{t+N\Delta}\}.
$
We define the masked regression loss as:
\begin{equation}
\mathcal{L}_{\mathrm{reg}}(\theta)
=
\frac{1}{\max(1, \sum_{n=1}^{N} m_{t+n\Delta})}
\sum_{n=1}^{N}
m_{t+n\Delta}
\left\|
\hat{y}_{t+n\Delta} - y_{t+n\Delta}
\right\|_1.
\label{eq:loss_trace_reg}
\end{equation}
We define the validity loss with binary cross-entropy:
\begin{equation}
\mathcal{L}_{\mathrm{valid}}(\theta)
=
\frac{1}{N}
\sum_{n=1}^{N}
\mathrm{BCE}
\left(
\hat{m}_{t+n\Delta}, m_{t+n\Delta}
\right).
\label{eq:loss_trace_valid}
\end{equation}
The auxiliary visual trace loss combines the regression and validity losses:
\begin{equation}
\mathcal{L}_{\mathrm{trace}}(\theta)
=
\mathcal{L}_{\mathrm{reg}}(\theta)
+
\lambda_{\mathrm{valid}}
\mathcal{L}_{\mathrm{valid}}(\theta),
\label{eq:loss_trace}
\end{equation}
where $\lambda_{\mathrm{valid}}$ controls the relative weight of the validity loss.
This loss is applied independently to each camera view and averaged across views.

\myparagraph{Overall training objective.}
Our final training objective combines the action prediction loss and the visual trace loss:
\begin{equation}
\mathcal{L}_{\mathrm{total}}(\theta)
=
\mathcal{L}_{\mathrm{act}}(\theta)
+
\lambda_{\mathrm{trace}} \mathcal{L}_{\mathrm{trace}}(\theta),
\label{eq:loss_total}
\end{equation}
where $\lambda_{\mathrm{trace}}$ controls the strength of the visual trace supervision.

\myparagraph{Inference.}
During inference, the VLM backbone processes the camera views, language instruction, and demonstration video to obtain the representations $\phi_t$, $\psi_t$, and $z_t$ at control timestep $t$.
The action expert predicts actions from $\phi_t$, $\psi_t$, $z_t$, $q_t$, and the current noisy action estimate.
We initialize $A_t^0 = \epsilon \sim \mathcal{N}(\mathbf{0}, \mathbf{I})$ and update it with Euler integration: $
A_t^{(k+1)/K}
=
A_t^{k/K}
+
\frac{1}{K}V_{\theta}(\phi_t, \psi_t, z_t, A_t^{k/K}, q_t)
$
for $k=0,\dots,K-1$.
After $K$ denoising steps, we obtain $A_t^1$ as the final predicted action chunk.

\section{RoboCasa-DC}
\label{sec:robocasa-dc}

\myparagraph{Base environment.}
RoboCasa~\citep{RoboCasa} provides a suite of 24 kitchen manipulation tasks with a 7-DoF Panda arm.
Each observation consists of proprioceptive states and three camera views, including two static cameras and one wrist-mounted camera.
The original benchmark provides approximately 3{,}000 expert Panda-arm trajectories per task for training, and evaluates trained policies on unseen seeds that vary scene layouts, object categories, and object placements.

\begin{figure}[t]
  \centering
  \includegraphics[width=\textwidth,trim=5 164 39 3,clip]{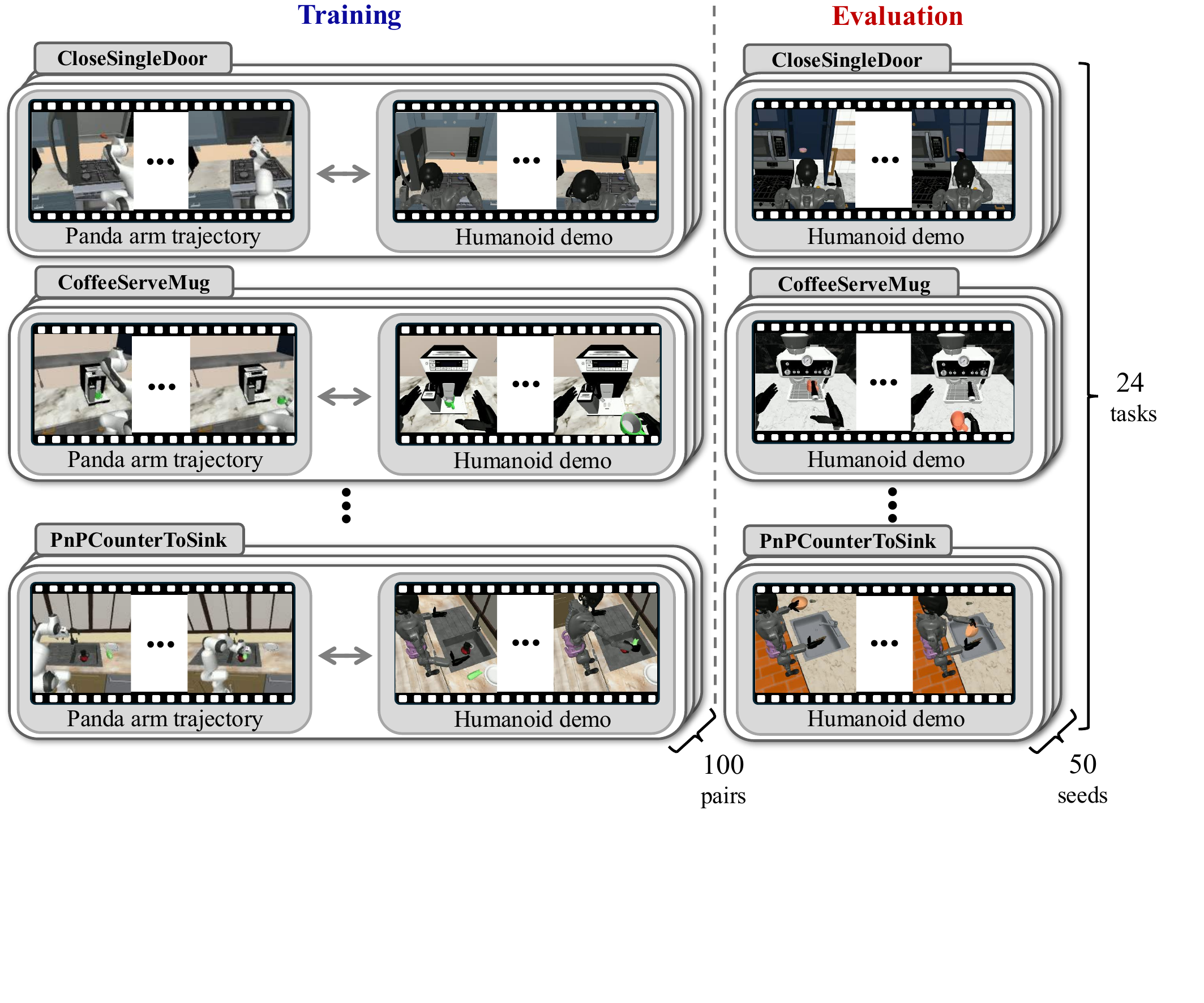}
  \caption{
  Cross-embodiment benchmark dataset in \textsc{RoboCasa-DC}.
  For each of the 24 tasks, we pair 100 original Panda-arm trajectories with collected GR-1 humanoid demonstrations for training.
  For evaluation, we collect humanoid demonstrations for 50 pre-defined seeds per task.
}

  \label{fig:robocasa-dc}
  \vskip -0.1in
\end{figure}

\myparagraph{Benchmark construction.}
\textsc{RoboCasa-DC} extends RoboCasa with demonstration videos that share the task and initial scene state with each Panda-arm trajectory, but may differ in embodiment and trajectory, allowing policies to condition on either cross- or same-embodiment demonstrations.
Fig.~\ref{fig:robocasa-dc} illustrates the cross-embodiment case, where Panda-arm trajectories are paired with corresponding GR-1 humanoid demonstrations.
For this setting, we pair 100 Panda-arm trajectories per task with corresponding humanoid expert demonstrations across all 24 tasks.
We collect each humanoid demonstration by restoring the corresponding Panda-arm trajectory's initial state and teleoperating the humanoid in the same scene with a Leap Motion controller, providing a simulated proxy for real-world human demonstrations.
For evaluation, we pre-define 50 evaluation seeds per task and collect corresponding humanoid demonstrations for each seed.
For each task, we manually select the demonstration camera view that best captures task execution.
We also include a same-embodiment case, following prior work~\citep{Uniskill, ViVLA}, where videos from Panda-arm trajectories are used as demonstrations.
We publicly release \textsc{RoboCasa-DC} to support reproducible evaluation and future work on demo-conditioned policies.

\myparagraph{Evaluation protocol.}
A \textsc{RoboCasa-DC} evaluation split partitions the 24 tasks into seen training tasks and unseen evaluation tasks.
Demo-conditioned policies are trained on the paired dataset from the seen tasks.
At test time, each policy is evaluated on evaluation seeds from unseen tasks while conditioning on the single demonstration video collected for each seed, which shares the same initial scene state.
This protocol tests whether a policy can extract task-relevant guidance from one-shot demonstrations, including cross-embodiment ones, rather than merely imitating expert trajectories from its own embodiment.

\section{Experiments}
\label{sec:exp}

\subsection{Experimental setup}

\myparagraph{RoboCasa-DC.}
We evaluate methods on two task splits of \textsc{RoboCasa-DC}: a \emph{category-balanced} split (similar to \citep{Meta-World}) and a \emph{precision-sensitive} split (motivated by \citep{RVT-2, PartInstruct}).
Each split holds out five tasks as unseen evaluation tasks, while the remaining 19 tasks are used for training.
The category-balanced split samples unseen tasks from different action categories:
\textit{CloseDrawer}, \textit{TurnOffSinkFaucet}, \textit{OpenDoubleDoor}, \textit{CoffeeServeMug}, and \textit{PnPCounterToMicrowave}.
The precision-sensitive split instead holds out five tasks with small target interaction ratio (TIR) (see Fig.~\ref{fig:tile}), which stress precise spatial grounding:
\textit{TurnOffStove}, \textit{CoffeePressButton}, \textit{TurnOffSinkFaucet}, \textit{PnPCounterToSink}, and \textit{PnPStoveToCounter}.
We use the dataset and evaluation protocol described in \S\ref{sec:robocasa-dc}.
For \textsc{SeeTraceAct}, the trace labels are generated automatically by projecting end-effector positions into each static camera view.

\myparagraph{Real-world benchmark.}
We evaluate demo-conditioned policies on real-world tabletop manipulation tasks using a Franka Panda arm (see Fig.~\ref{fig:real:tasks} in App.~\ref{sec:bench-det:real}).
For each of four seen tasks---\textit{Pick Coke}, \textit{Stack Blocks}, \textit{Stack Cups}, and \textit{Close Laptop}---we collect 150 episode pairs, each consisting of a teleoperated Franka trajectory and a human demonstration performed from the same initial scene;
policies are trained on these pairs.
For \textsc{SeeTraceAct}, we obtain trace labels from the robot trajectories during collection by projecting end-effector positions into the static camera view, following \citep{RT-Trajectory,HAMSTER}.
For evaluation, we define six unseen tasks: \textit{Pick Block}, \textit{Pick Cup}, \textit{Press Button}, \textit{Stack Blocks in Swapped Order}, \textit{Place Block in Bin}, and \textit{Close Kettle Lid}.
These tasks are designed by recombining seen skills and objects, transferring seen skills to novel objects, and introducing a new contact interaction.
We test each unseen task across 10 different object positions and poses.
Following prior evaluation setups~\citep{Uniskill, ViVLA, xskill}, each policy is conditioned on a one-shot human demonstration video captured in a closely matched scene.
We report per-task and average success rates over the unseen tasks, further averaged over three independent runs.
Hardware setup and task language instructions are provided in Fig.~\ref{fig:real:setup} and Table~\ref{tab:real:lang}, respectively.

\myparagraph{Baselines.}
We compare \textsc{SeeTraceAct} against three demo-conditioned baselines: Vid2Robot~\citep{Vid2Robot}, UniSkill~\citep{Uniskill}, and ViVLA~\citep{ViVLA}.
For fair comparison, all baselines are re-implemented on top of the same GR00T N1.5 backbone as \textsc{SeeTraceAct} and trained with their corresponding objectives.
Further implementation details are provided in App.~\ref{sec:model-det}.

\subsection{Results}
\label{sec:exp:results}

Table~\ref{tab:sim} summarizes the experimental results on \textsc{RoboCasa-DC}.
\textsc{SeeTraceAct} achieves the best performance across all four evaluation settings, covering both category-balanced and precision-sensitive splits under same-embodiment and cross-embodiment demonstrations.
This consistency is important because the two axes stress different aspects of demo-conditioned control: the precision-sensitive split requires fine-grained spatial grounding, while the cross-embodiment setting requires extracting task-relevant motion from demonstrations whose embodiment differs from the robot.
Although absolute performance generally decreases under these harder settings, \textsc{SeeTraceAct} maintains its advantage over the baselines, with the largest margin appearing in the precision-sensitive cross-embodiment setting.

\begin{figure}[t]
	\centering
	\includegraphics[width=\textwidth,trim=7 7 10 12,clip]{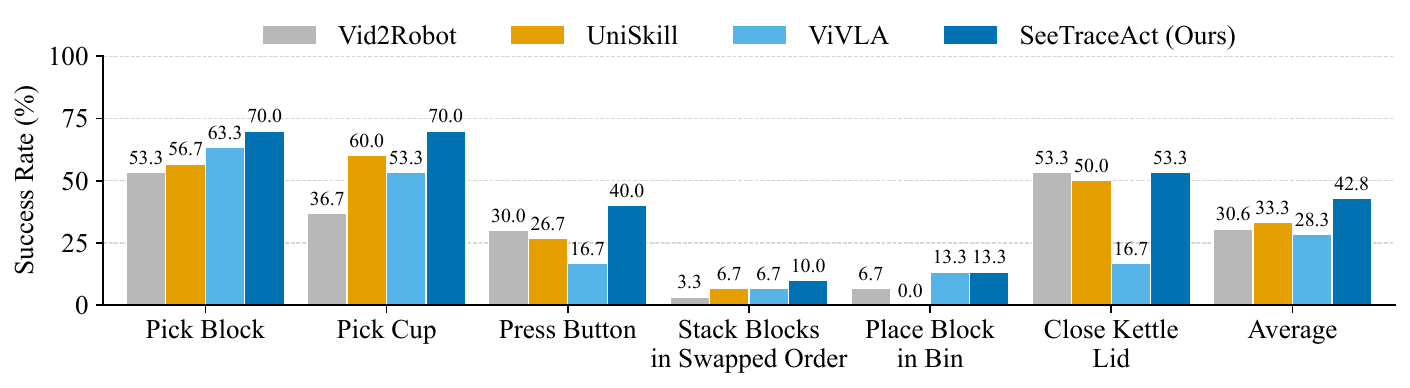}
	\caption{Experimental results on the real-world benchmark with a Franka Panda arm.
  We report success rates over 10 trials for each of the six unseen tasks, averaged over three independent runs, along with the average across tasks.}
	\label{fig:real:results}
\end{figure}

Fig.~\ref{fig:real:results} shows the experimental results on the real-world benchmark.
\textsc{SeeTraceAct} again matches or outperforms every baseline on all six unseen tasks, improving average success from 33.3\% to 42.8\% over the strongest baseline.
These results extend the gains observed in simulation to real-world cross-embodiment demo conditioning.

\subsection{Larger gains on precision-sensitive tasks}
\label{sec:exp:contrast}

Table~\ref{tab:sim} shows that \textsc{SeeTraceAct}'s margin over the strongest baseline is larger on the precision-sensitive split than on the category-balanced split, increasing from 0.7 to 2.6 percentage points on average across the same- and cross-embodiment settings.
This suggests that visibility-aware trace supervision is especially useful when successful execution depends on accurately localizing a small task-critical interaction region.

To examine this trend more directly, we train all models on all 24 \textsc{RoboCasa-DC} tasks and evaluate them on 50 held-out seeds per task (Table~\ref{tab:all-tasks-same-emb}), similar to \citep{RVT-2}.
We use the same-embodiment demonstrations to factor out the cross-embodiment gap.
In this setting, \textsc{SeeTraceAct} again achieves the highest average success rate.
To characterize where these gains arise, we measure precision sensitivity using the target interaction ratio (TIR), defined as the fraction of the image covered by the target interaction region (Fig.~\ref{fig:tile}); a smaller TIR indicates a more precision-sensitive task.
We compare TIR with the per-task improvement of \textsc{SeeTraceAct} over each baseline using a one-sided Spearman rank correlation test.
Across the 24 tasks, TIR is negatively correlated with the gain over every baseline: Vid2Robot ($\rho=-0.80$, $p<10^{-5}$), UniSkill ($\rho=-0.45$, $p<0.05$), and ViVLA ($\rho=-0.52$, $p<0.01$).
The correlation remains significant even when we compare \textsc{SeeTraceAct} against the strongest baseline for each task ($\rho=-0.63$, $p<10^{-3}$), consistent with the larger gains on precision-sensitive tasks in Table~\ref{tab:sim}.

\begin{table}[t]
  \begin{center}
    \begin{small}
        \begin{tabular}{lcccc}
          \toprule
          & \multicolumn{2}{c}{Category-balanced split} & \multicolumn{2}{c}{Precision-sensitive split} \\
          \cmidrule(lr){2-3} \cmidrule(lr){4-5}
          Method  & Same-emb demo & Cross-emb demo & Same-emb demo & Cross-emb demo \\
          \midrule
          Vid2Robot \citep{Vid2Robot} & 17.9\% & 9.5\% & 11.4\% & 5.6\% \\
          UniSkill \citep{Uniskill} & 13.0\% & 11.3\% & 11.7\% & 5.6\% \\
          ViVLA \citep{ViVLA} & 14.1\% & 9.1\% & 8.8\% & 7.2\% \\
          \midrule
          \textsc{SeeTraceAct} (Ours) & \textbf{18.9\%} & \textbf{11.6\%} & \textbf{13.2\%} & \textbf{10.8\%} \\
          \bottomrule
        \end{tabular}
    \end{small}
  \end{center}
  \caption{
    Experimental results on \textsc{RoboCasa-DC}. We report success rates averaged over the five unseen tasks in each split, with 50 evaluation episodes per task, and further averaged over three independent runs.
    In \emph{Same-emb demo}, policies are conditioned on demonstrations from the same Panda-arm embodiment; in \emph{Cross-emb demo}, policies are conditioned on simulated GR-1 humanoid demonstrations as a proxy for human videos.
  }
  \label{tab:sim}
  % \vskip -0.15in
\end{table}

\subsection{Ablation study}
\label{sec:exp:ablation}

\begin{table}[h]
  \begin{center}
    \begin{small}
        \begin{tabular}{lc}
          \toprule
          Method  & Success rate \\
          \midrule
          GR00T N1.5 backbone (no demonstrations) & 8.5\% \\
          GR00T N1.5 backbone w/ trace supervision (no demonstrations) & 9.0\% \\
          \textsc{SeeTraceAct} w/o trace supervision & 9.3\% \\
          \textsc{SeeTraceAct} w/o validity head & 8.6\% \\
          \textsc{SeeTraceAct} w/o action-aware video encoder & 8.1\% \\
          \textsc{SeeTraceAct} w/ 3D trace supervision & 9.7\% \\
          \textsc{SeeTraceAct} w/ demonstration trace supervision & 8.7\% \\
          \midrule
          \textsc{SeeTraceAct} (Ours) & \textbf{11.2\%} \\
          \bottomrule
        \end{tabular}
    \end{small}
  \end{center}
  \caption{Ablation study in the cross-embodiment setting of \textsc{RoboCasa-DC}.
  Success rates are averaged over the category-balanced and precision-sensitive splits, and further averaged over three independent runs.}
  \label{tab:ablation}
  \vskip -0.15in
\end{table}

We ablate the design choices of \textsc{SeeTraceAct} in the cross-embodiment setting of \textsc{RoboCasa-DC}, averaging results across the category-balanced and precision-sensitive splits.
We also report the GR00T N1.5 backbone without demonstrations, with and without our trace supervision, as reference points for standard VLA policies.
Table~\ref{tab:ablation} shows that neither demonstration inputs nor trace supervision is sufficient on its own.
Removing trace supervision from \textsc{SeeTraceAct} reduces performance to 9.3\%, while adding trace supervision to the no-demonstration backbone yields only 9.0\%.
Removing the validity head lowers performance to 8.6\%, below even the variant without any trace supervision.
This supports the importance of visibility-aware supervision; forcing the model to regress ill-defined off-screen coordinates can hurt representation learning.
Replacing the V-JEPA 2~\citep{V-JEPA-2} video encoder with the backbone VLA's SigLIP~\citep{SigLIP} encoder also lowers performance to 8.1\%, suggesting that an action-aware video representation is useful for extracting task progression from demonstrations.
We further consider two alternative trace targets.
For 3D trace supervision, we append to each 2D point the metric depth along the viewing direction, computed from the logged end-effector positions and camera calibration.
This variant remains the strongest ablation but still lags behind image-space supervision, likely because 2D traces across multiple views already constrain the 3D position, so adding depth offers little new information while introducing a target that the RGB-pretrained backbone must infer implicitly.
Finally, supervising traces of the demonstrator's end effector in the demonstration video, instead of the robot's, lowers performance to 8.7\%.

\section{Limitations}

Our study has several limitations.
First, our real-world evaluation is limited to a tabletop single-arm benchmark, leaving broader evaluation across embodiments, environments, and tasks for future work.
Second, our auxiliary objective assumes future trace labels during training; while they can be computed from end-effector positions and camera calibration already provided by many robot datasets~\citep{DROID}, when these are unavailable, such labels could be obtained using motion trackers or foundation models, following prior work~\citep{LLARVA,TraceVLA,ThinkAct,FastThinkAct,MolmoAct,RT-Trajectory}.
Finally, despite improving over baselines, the modest absolute success rates on \textsc{RoboCasa-DC} show that demo-conditioned robot learning leaves substantial room for progress.

\section{Conclusion}

We presented \textsc{SeeTraceAct}, a demo-conditioned VLA framework that grounds one-shot demonstrations through visibility-aware latent planning.
We also introduced \textsc{RoboCasa-DC}, a benchmark for evaluating demo-conditioned policies with same- and cross-embodiment demonstrations.
Experiments show that \textsc{SeeTraceAct} achieves the strongest performance among baselines, with especially large gains on precision-sensitive tasks.
These results highlight precise spatial grounding as a promising direction for demo-conditioned robot learning.

\clearpage
% The acknowledgments are automatically included only in the final and preprint versions of the paper.

\acknowledgments{We thank Jaeah Lee, Hanjung Kim, and the anonymous reviewers for their insightful feedback, and Junghwan Yim for data collection in the course project that preceded this work. This material is based upon work partially supported by the National Science Foundation under Grant No. 2239292 and NVIDIA Academic Grant Program.}

%===============================================================================

\bibliography{vla-video}  % .bib

\newpage
\appendix
\onecolumn
\section{Benchmark Details}
\label{sec:bench-det}

\subsection{RoboCasa-DC}
\label{sec:bench-det:robocasa-dc}

To collect GR-1 humanoid demonstrations, we restore the initial simulation state of each pre-defined Panda-arm trajectory and teleoperate the humanoid in the same scene.
Because the released RoboCasa trajectories do not include the generative fixture textures needed to exactly restore the original rendered observations, we randomly assign fixture textures, replay the Panda-arm trajectories, and re-render the corresponding videos.
For the cross-embodiment setting, we use the paired Panda--GR-1 episodes described in \S\ref{sec:robocasa-dc} for training and the pre-defined evaluation seeds for testing.
For the same-embodiment case, demonstrations are rendered from the Panda-arm trajectories themselves, allowing us to use the full RoboCasa training set, up to approximately 3{,}000 trajectories per task.

\subsection{Real-world benchmark}
\label{sec:bench-det:real}

\begin{figure}[h]
  \centering
  \includegraphics[width=\textwidth,trim=0 270 0 25,clip]{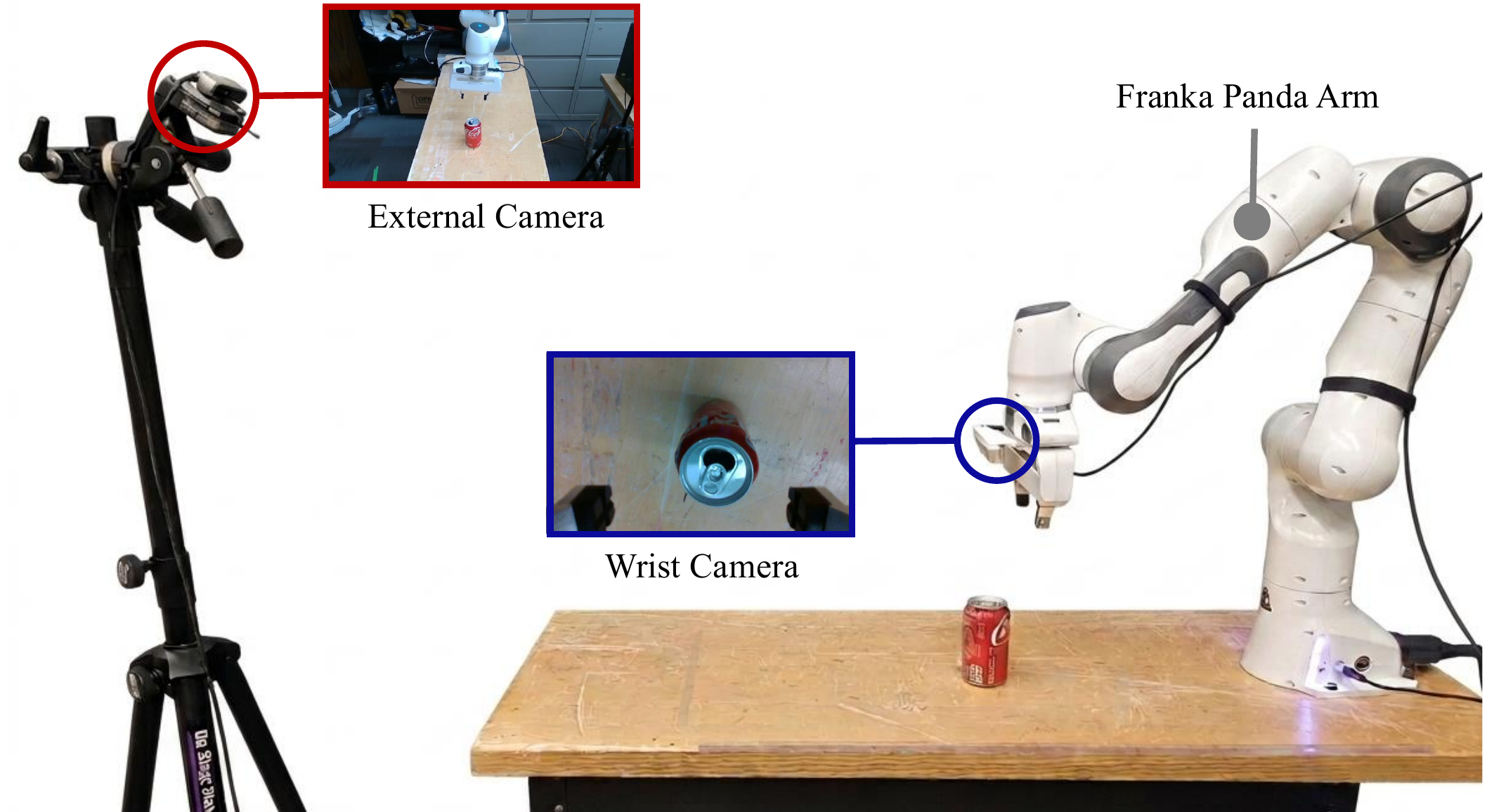}
  \caption{Hardware setup for the real-world experiments.
  The setup consists of a Franka Panda arm, an external third-person-view camera, and a wrist-mounted camera.}
  \label{fig:real:setup}
  \vskip -0.1in
\end{figure}

\myparagraph{Hardware setup.}
Our real-world experiments use a 7-DoF Franka Panda arm mounted on a tabletop and equipped with a parallel-jaw gripper (Fig.~\ref{fig:real:setup}).
The visual observations are captured from two RGB cameras: a single external camera providing the third-person view and a wrist-mounted camera attached near the end effector.
Both camera streams are captured at a resolution of $640 \times 480$.

\begin{figure}[h]
  \centering

  \begin{subfigure}[t]{\textwidth}
    \centering
    \includegraphics[width=\textwidth,trim=10 635 10 3,clip]{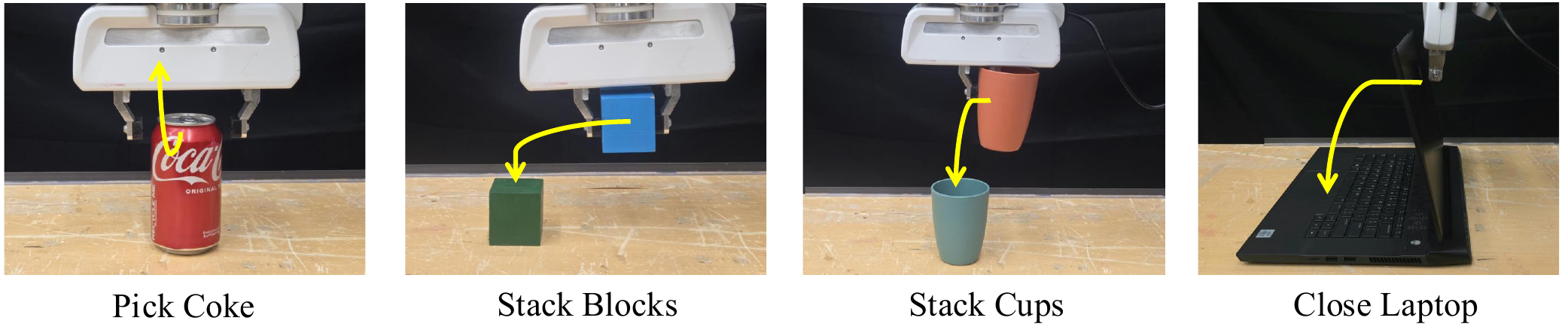}
    \caption{Seen tasks}
  \end{subfigure}

  \vspace{0.7em}
    \begin{subfigure}[t]{0.74\textwidth}
    \centering
    \includegraphics[width=\textwidth,trim=10 392 270 3,clip]{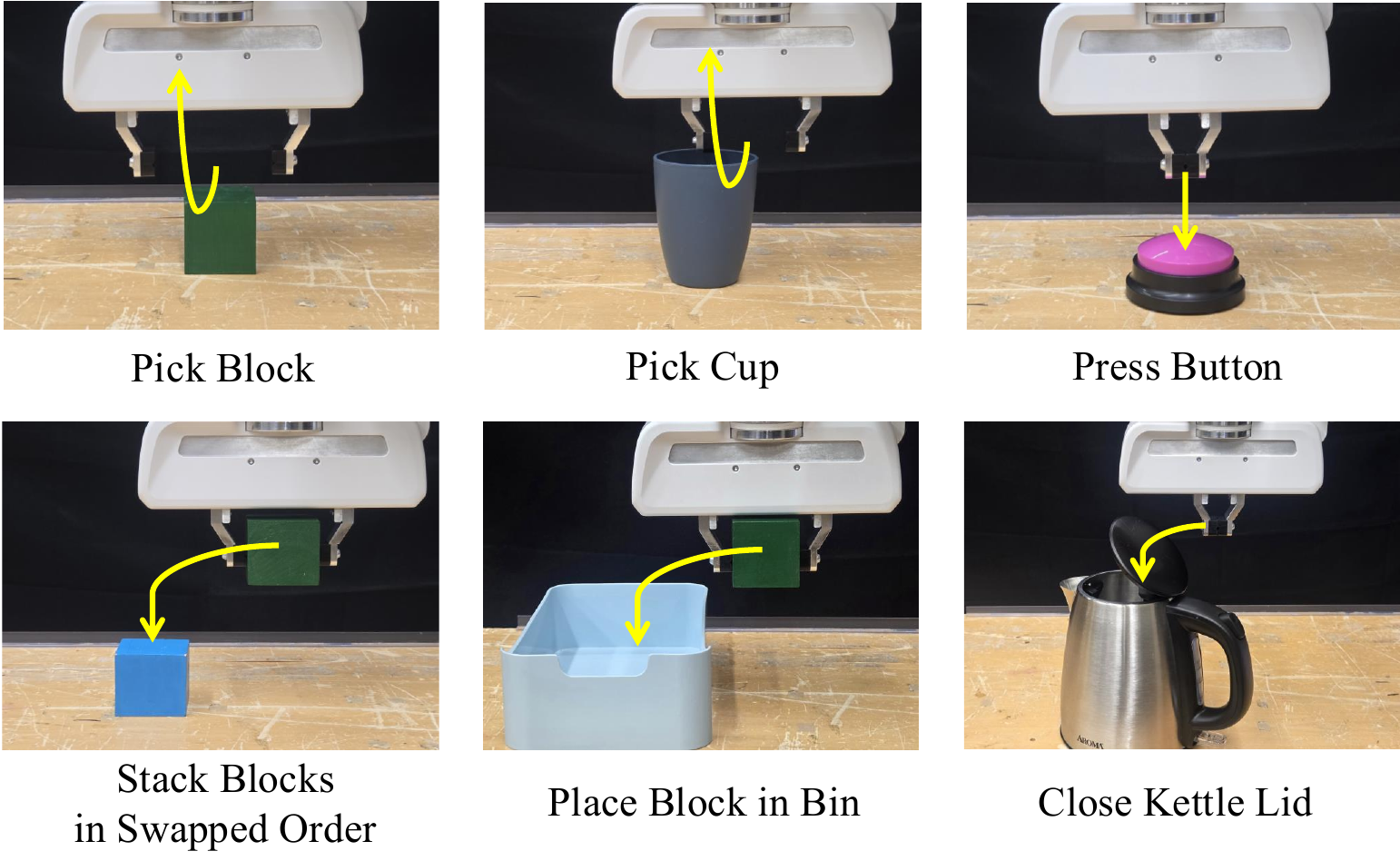}
    \vspace{-1.5em}
    \caption{Unseen tasks}
  \end{subfigure}

  \vspace{-0.2em}

  \caption{(a) Four seen tasks and (b) six unseen tasks in the real-world benchmark.
  The yellow arrow indicates the desired path of the end effector.}
  \label{fig:real:tasks}
  \vspace{-0.1in}
\end{figure}

\myparagraph{Tasks.}
The language instructions for the real-world tasks shown in Fig.~\ref{fig:real:tasks} are listed in Table~\ref{tab:real:lang}.

\begin{table}[h]
  \begin{center}
  \begin{small}
    \begin{tabular}{ll}
      \toprule
      Task & Language instruction \\
      \midrule
      \multicolumn{2}{l}{\textbf{Seen tasks}} \\
      Pick Coke & ``Grab the coke can and lift it up'' \\
      Stack Blocks & ``Stack the blue block on top of the green block'' \\
      Stack Cups & ``Stack the orange cup onto the green cup'' \\
      Close Laptop & ``Close the laptop lid fully'' \\
      \midrule
      \multicolumn{2}{l}{\textbf{Unseen tasks}} \\
      Pick Block & ``Grab the green block and lift it up'' \\
      Pick Cup & ``Grab the blue cup and lift it up'' \\
      Press Button & ``Press the pink button'' \\
      Stack Blocks in Swapped Order & ``Stack the green block on top of the blue block'' \\
      Place Block in Bin & ``Place the block into the bin'' \\
      Close Kettle Lid & ``Close the kettle lid fully'' \\
      \bottomrule
    \end{tabular}
    \end{small}
  \end{center}
  \caption{Real-world benchmark tasks and corresponding language instructions.} 
  \label{tab:real:lang}
  \vskip -0.1in
\end{table}

\section{Model Details}
\label{sec:model-det}

\begin{table}[t]
  \begin{center}
  \begin{small}
  \begin{tabular}{lcccc}
    \toprule
    Hyperparameter & Vid2Robot & UniSkill & ViVLA & SeeTraceAct \\
    \midrule
    \multicolumn{5}{l}{\textbf{Training}} \\
    Training steps & \multicolumn{4}{c}{50{,}000} \\
    Batch size & \multicolumn{4}{c}{128} \\
    Optimizer & \multicolumn{4}{c}{AdamW ($\beta_1{=}0.95$, $\beta_2{=}0.999$, $\epsilon{=}10^{-8}$)} \\
    Learning rate & \multicolumn{4}{c}{$10^{-4}$} \\
    Weight decay & \multicolumn{4}{c}{$10^{-5}$} \\
    Warmup steps & \multicolumn{4}{c}{2500} \\
    Learning rate scheduler & \multicolumn{4}{c}{cosine} \\
    Gradient clip norm & \multicolumn{4}{c}{1.0} \\
    \midrule
    \multicolumn{5}{l}{\textbf{Demonstration video}} \\
    Encoder architecture & SigLIP + Perceiver & ISD & \multicolumn{2}{c}{V-JEPA 2 + Perceiver} \\
    Frame sampling method & uniform & timestep & \multicolumn{2}{c}{uniform} \\
    Encoder input resolution & 224$\times$224 & 224$\times$224 & \multicolumn{2}{c}{256$\times$256} \\
    \# frames sampled & 64 & --- & \multicolumn{2}{c}{64} \\
    \# video tokens & 32 & 1 & \multicolumn{2}{c}{32} \\
    \midrule
    \multicolumn{5}{l}{\textbf{Action prediction}} \\
    Action horizon ($H$) & \multicolumn{4}{c}{16} \\
    Denoising steps ($K$) & \multicolumn{4}{c}{4} \\
    \midrule
    \multicolumn{5}{l}{\textbf{Trace supervision}} \\
    Validity loss weight ($\lambda_{\mathrm{valid}}$) & --- & --- & --- & 1.0 \\
    Visual trace loss weight ($\lambda_{\mathrm{trace}}$) & --- & --- & --- & 0.2 \\
    \# future trace points ($N$) & --- & --- & --- & 5 \\
    Temporal stride ($\Delta$) & --- & --- & --- & 5 \\
    \bottomrule
  \end{tabular}
  \end{small}
\end{center}
  \caption{Hyperparameters used for the compared methods on \textsc{RoboCasa-DC}.
  \textit{Encoder input resolution} follows each video encoder's default.
  Following~\citep{Uniskill}, UniSkill extracts demonstration frames at timesteps $t$ and $t+k$ at each control timestep, where $k$ is the skill interval, while the other methods use uniformly sampled frames from the episode.}
  \label{tab:hyp}
  \vskip -0.1in
\end{table}

\myparagraph{Common setup.}
For fair comparison, all methods are implemented on top of the same GR00T N1.5 architecture.
Each method receives the same camera views, language instruction, demonstration video, and robot states as input, and predicts action chunks with the same horizon.
Each baseline method is trained with the corresponding objectives proposed in the original work.
For action prediction, we use an action horizon of $H=16$ and perform $K=4$ denoising steps at inference time.
For the action prediction loss, we sample $u \sim \mathrm{Beta}(1.5, 1.0)$ and set $\tau=(s-u)/s$ with $s=0.999$, following prior work~\citep{pi0,GR00T}.
The hyperparameters used in our experiments are summarized in Table~\ref{tab:hyp}.
Further details are provided in our code.

\myparagraph{SeeTraceAct.}
For demonstration video encoding, we follow the open-sourced V-JEPA 2 preprocessing and uniformly sample 64 frames from each demonstration video.
The sampled frames are resized to the video encoder input resolution of $256 \times 256$.
Directly feeding all video features into the VLM would substantially increase the input sequence length.
With a 3D patch size of $2 \times 16 \times 16$, a 64-frame video produces $32 \times 16 \times 16 = 8192$ video tokens, which is much larger than the approximately 200 image and language tokens used by the backbone.
We therefore employ a Perceiver Resampler to compress the video features into 32 video tokens before appending them to the VLM input sequence.
As in the original GR00T N1.5, robot observation images are encoded with its SigLIP~\citep{SigLIP} image encoder.

\begin{table}[h]
  \begin{center}
  \begin{small}
    \begin{tabular}{lccccc}
      \toprule
      Task & Vid2Robot & UniSkill & ViVLA & SeeTraceAct & TIR \\
      \midrule
      CloseDoubleDoor        & 83.3  & 51.9  & 85.2 & 72.2 & 6.403 \\
      CloseDrawer            & 98.1  & 100.0 & 94.4 & 94.4 & 4.736 \\
      CloseSingleDoor        & 94.4  & 96.3  & 94.4 & 92.6 & 8.801 \\
      CoffeePressButton      & 64.8  & 37.0  & 70.4 & 83.3 & 0.092 \\
      CoffeeServeMug         & 53.7  & 57.4  & 74.1 & 53.7 & 0.507 \\
      CoffeeSetupMug         & 25.9  & 18.5  & 27.8 & 24.1 & 0.708 \\
      OpenDoubleDoor         & 20.8  & 13.2  & 41.5 & 17.0 & 1.349 \\
      OpenDrawer             & 33.3  & 48.1  & 53.7 & 48.1 & 0.281 \\
      OpenSingleDoor         & 57.4  & 29.6  & 55.6 & 61.1 & 0.439 \\
      PnPCabToCounter        & 42.6  & 51.9  & 44.4 & 50.0 & 0.427 \\
      PnPCounterToCab        & 33.3  & 11.1  & 22.2 & 33.3 & 0.671 \\
      PnPCounterToMicrowave  & 20.4  & 14.8  & 22.2 & 18.5 & 0.452 \\
      PnPCounterToSink       & 37.0  & 14.8  & 31.5 & 38.9 & 0.153 \\
      PnPCounterToStove      & 30.2  & 7.5   & 32.1 & 26.4 & 0.366 \\
      PnPMicrowaveToCounter  & 27.8  & 9.3   & 51.9 & 33.3 & 0.275 \\
      PnPSinkToCounter       & 7.4   & 7.4   & 18.5 & 22.2 & 0.165 \\
      PnPStoveToCounter      & 40.7  & 18.5  & 51.9 & 48.1 & 0.165 \\
      TurnOffMicrowave       & 79.6  & 48.1  & 74.1 & 83.3 & 0.262 \\
      TurnOffSinkFaucet      & 75.9  & 72.2  & 75.9 & 85.2 & 0.116 \\
      TurnOffStove           & 37.0  & 42.6  & 46.3 & 50.0 & 0.049 \\
      TurnOnMicrowave        & 64.8  & 63.0  & 68.5 & 66.7 & 0.470 \\
      TurnOnSinkFaucet       & 20.4  & 37.0  & 31.5 & 46.3 & 0.189 \\
      TurnOnStove            & 60.4  & 66.0  & 54.7 & 79.2 & 0.067 \\
      TurnSinkSpout          & 87.0  & 94.4  & 88.9 & 87.0 & 0.336 \\
      \midrule
      Average                & 49.9  & 42.1  & 54.7 & 54.8 & 1.139 \\
      \bottomrule
    \end{tabular}
    \end{small}
  \end{center}
  \caption{Per-task success rates when training all methods on all 24 \textsc{RoboCasa-DC} tasks.
  Results are reported in the same-embodiment setting, with evaluation on 50 held-out seeds per task.
  We also report the target interaction ratio (TIR), defined as the area ratio of the target interaction region to the full camera view, as visualized in Fig.~\ref{fig:tile}.
  Lower TIR indicates higher precision sensitivity.
  Success rates and TIR are reported as percentages.}
  \label{tab:all-tasks-same-emb}
  \vskip -0.1in
\end{table}

\myparagraph{Vid2Robot~\citep{Vid2Robot}.}
Since Vid2Robot does not provide an official implementation, we re-implement the method described in the paper within our common GR00T N1.5 backbone.
The original method uses a shared visual encoder for robot observations and prompt videos, so we use the SigLIP encoder from GR00T N1.5 to encode demonstration videos as well.
We then pass the resulting video features through a Perceiver Resampler to compress them into 32 video tokens.
We train the model with the action prediction loss, the prompt--robot video contrastive loss (VVCL), and the video--text contrastive loss (VTCL).
In the same-embodiment setting, we omit VVCL because it already attains its theoretical minimum.

\myparagraph{UniSkill~\citep{Uniskill}.}
We use the inverse skill dynamics (ISD) module from the official UniSkill implementation.
At each control timestep $t$, UniSkill passes the demonstration frames at $t$ and $t+k$ to the ISD module, where $k$ is the skill interval, to obtain a skill representation for subsequent action prediction.
Following the original implementation, we set $k=20$ frames.
We project the resulting skill representation into the VLM embedding space and append it as a single token to the input image, language, and video tokens.

\myparagraph{ViVLA~\citep{ViVLA}.}
Because ViVLA does not provide an official implementation, we re-implement the method described in the paper within our common GR00T N1.5 backbone.
To isolate the effect of the latent action prediction objective rather than the video encoder, we use the same V-JEPA 2 video encoder and Perceiver Resampler as those used in \textsc{SeeTraceAct} to generate input video tokens.
For the latent action encoder, we use the open-sourced LAPA~\citep{LAPA}.
We remove the prediction of the number of LACT tokens and instead fix the number of tokens to 9, allowing ViVLA to generate each action chunk with a single forward pass.
Specifically, we divide the demonstration video into 8 equal segments and train 8 LACT tokens to predict the LAPA representations for the corresponding segments, while the remaining LACT token predicts the representation of the robot's subsequent latent action.
These 9 auxiliary tokens provide at least as much capacity as the 5 learnable query tokens used by \textsc{SeeTraceAct}.

\begin{figure}[h]
  \centering
  \includegraphics[width=\textwidth,trim=0 0 0 0,clip]{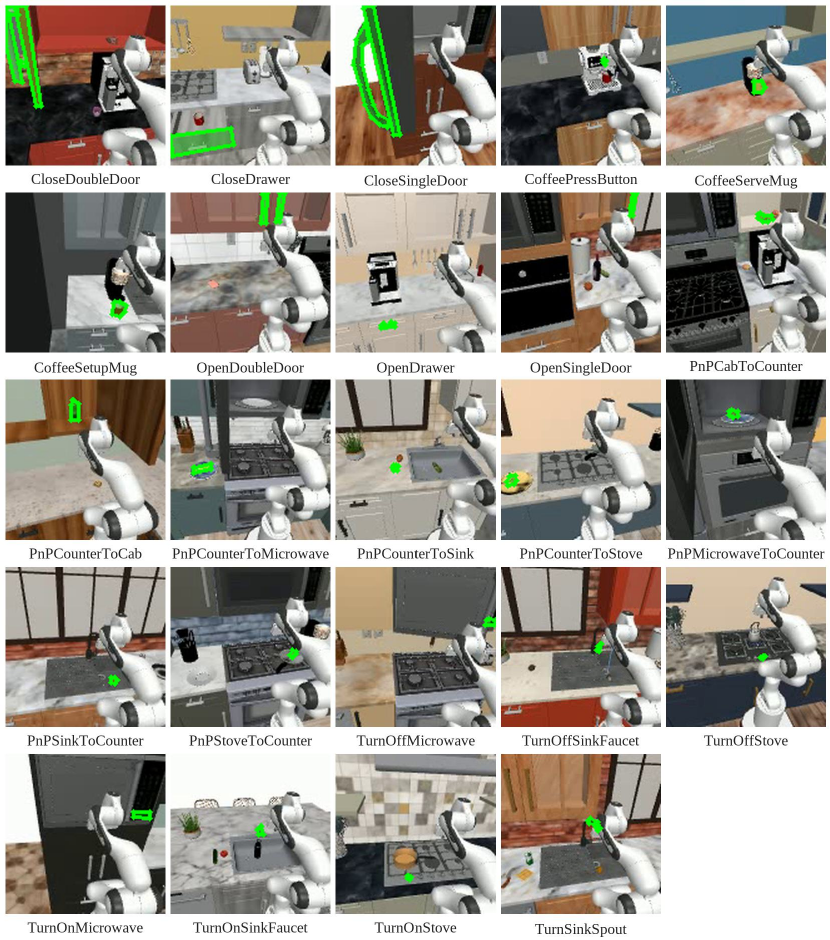}
  \caption{Target interaction regions in \textsc{RoboCasa-DC} tasks. The highlighted region indicates the area in a static camera view where the robot must interact to complete the task. We use the area ratio of this region to the full camera view as the \textit{target interaction ratio} (TIR) in \S\ref{sec:exp} and Table~\ref{tab:all-tasks-same-emb}; lower TIR indicates a more precision-sensitive task.}
  \label{fig:tile}
  \vskip -0.1in
\end{figure}

\end{document}